\documentclass[conference]{IEEEtran}
\IEEEoverridecommandlockouts
\usepackage{cite}
\usepackage{amsmath,amssymb,amsfonts}
\usepackage{algorithmic}
\usepackage{graphicx}
\usepackage{textcomp}
\usepackage{xcolor}
\usepackage{stfloats} 
\usepackage{pifont}               
\usepackage{multirow}
\usepackage{url}
\newcommand{\cmark}{\ding{51}}    
\newcommand{\xmark}{\ding{55}}    
\usepackage{booktabs}
\usepackage{xcolor}
\usepackage{graphicx}
\usepackage{makecell} 
\usepackage{array}    

\def\BibTeX{{\rm B\kern-.05em{\sc i\kern-.025em b}\kern-.08em
    T\kern-.1667em\lower.7ex\hbox{E}\kern-.125emX}}
\begin{document}

\title{Neural Video Compression with Domain Transfer
}

\author{
Tiange Zhang$^{*\dagger\ddagger}$\thanks{Tiange Zhang and Rongqun Lin contributed equally to this work. Corresponding authors: Siwei Ma (swma@pku.edu.cn) and Xiandong Meng (mengxd@pcl.ac.cn). This work was supported in part by the Key Research $\And$ Development Program of Pengcheng Laboratory under grant PCL2024A02 and in part by Natural Science Foundation of China No. 62371008, and New Cornerstone Science Foundation through the XPLORER PRIZE.}, 
Rongqun Lin$^{\dagger}$, 
Xiandong Meng$^{\dagger}$, 
Haofeng Wang$^{*}$, 
Xing Tian$^{\dagger}$, 
Qi Zhang$^{\dagger}$, 
Siwei Ma$^{*\dagger\ddagger}$\\
Email: \{tgzhang, hfwang\}@stu.pku.edu.cn, \{linrq, mengxd\}@pcl.ac.cn, txsing@live.com, \\qizhang@alumni.pku.edu.cn, swma@pku.edu.cn\\
$^{*}$Shenzhen Graduate School, Peking University, Shenzhen, China\\
$^{\dagger}$Pengcheng Laboratory, Shenzhen, China\\
$^{\ddagger}$School of Computer Science, Peking University, Beijing, China
}

\maketitle
\begin{abstract}
Content-adaptive compression has always been a key direction in neural video coding (NVC), aiming to mitigate the domain gap between training and testing data. Such gaps often arise from distributional discrepancies between training and inference data, which may cause noticeable performance degradation when the testing content differs from the training distribution. To tackle this challenge, we propose DCVC-DT, a domain transfer enhanced neural video compression framework. Specifically, we design a lightweight online domain transfer (DT) mechanism that dynamically adapts the encoded latent representation during inference, effectively bridging the domain gap without modifying the encoder or decoder parameters. In addition, we develop a frame-level dynamic RD (Rate and Distortion) adjustment scheme that actively regulates the ratio of $R$ and $D$ in the loss function based on quality fluctuation, thereby improving rate-distortion performance. Extensive experiments demonstrate that DCVC-DT achieves up to 6.21\% bitrate savings over the baseline DCVC-DC, while significantly enhancing generalization to unseen testing data and alleviating error propagation. Our code is available at \url{https://github.com/SunnyMass/DCVC-DT}.

\end{abstract}


\section{Introduction}

Content-adaptive neural video compression (NVC) has become a key direction in recent years, aiming to enhance the adaptability of pre-trained models across diverse video contents. Although various existing NVCs~\cite{li2023neural,10558613,liao2025efvc,sheng2025bi,zhai2025lbvc,jia2025emerging,ye2024deep,LIN2023126396,SIP-20250056,zhang2025content,zhang2025stec,meng2023learned,jiang2025ecvc,sheng2024spatial,zhai2024hybrid,sheng2025prediction,jiang2025biecvc,lin2022dmvc,liao2025dynamic,liao2024neural,feng2025staco,liao2024rate,11044111,11462194} have achieved remarkable progress and outperform traditional codecs~\cite{wiegand2003overview,sullivan2012overview,9503377}, their performance still relies heavily on the domain consistency between training and testing data. In practice, non-natural content such as screen recordings, animations, and medical videos often exhibit distinct statistical properties from natural videos, resulting in severe domain gaps that degrade both coding efficiency and generalization ability.

In deep image compression, significant efforts have been made to address domain differences. Initially, Campos et al.~\cite{campos2019content} refined latent representations via an iterative procedure while keeping network parameters fixed. This strategy was further advanced by Yang et al.~\cite{yang2020improving} through optimized inference to address gaps in amortization, discretization, and marginalization. Beyond latent updates, other approaches involve parameter fine-tuning or incorporating adaptive modules. For instance, Van Rozendaal et al.~\cite{van2021overfitting} refined the entire network architecture, while Tsubota et al.~\cite{tsubota2023universal} proposed a universal framework combining latent refinement with lightweight decoder adapters. Similarly, Pan et al.~\cite{pan2022content} utilized adaptive strategies like content-adaptive channel dropping and Spatial Feature Transformation (SFT) layers. These advancements demonstrated the potential of combining parameter optimization with adaptive mechanisms to enhance compression performance.

Similar to image compression, recent video compression methods also explored content-adaptive strategies. Tang et al.~\cite{tang2024offline} enhanced optical flow by fine-tuning a pre-trained network with motion information from VTM in an offline stage, followed by online optimization of latent features. However, this primarily improves motion estimation. Chen et al.~\cite{chen2024group} applied parameter-efficient transfer learning (PETL) through low-rank adapters and GoP-level optimization, but still requires partial parameter updates and increases encoder complexity. These approaches primarily focus on the motion coding components, overlooking the latent fine-tuning of context coding modules, which account for a larger portion of the bitrate. Moreover, the current parameter fine-tuning methods introduce high complexity in video compression models.

Motivated by these observations, we propose DCVC-DT, a domain transfer enhanced neural video compression framework that achieves content adaptation purely in the latent domain. Instead of updating network parameters or introducing additional modules, our method dynamically adjusts the encoded latent representation on the encoder side through an online domain transfer (DT) mechanism, enabling efficient adaptation to unseen domains while maintaining decoding time. Furthermore, a frame-level dynamic RD adjustment strategy guided by quality fluctuation is designed to actively regulate the ratio of $R$ and $D$ in the loss function for better rate–distortion performance.

In summary, our main contributions are as follows:
\begin{itemize}
    \item We introduce a lightweight online domain transfer mechanism that performs adaptive latent refinement during inference, effectively bridging the domain gap without modifying model parameters or decoder architecture.
    \item We develop a frame-level dynamic RD adjustment strategy that actively regulates the ratio of $R$ and $D$ in the loss function based on quality fluctuation, improving rate–distortion performance.
    \item Extensive experiments demonstrate that DCVC-DT achieves up to 6.21\% bitrate savings over the baseline DCVC-DC and maintains strong cross-domain generalization across heterogeneous video datasets.

\end{itemize}

\begin{figure*}[!hbt]
    \centering
    \includegraphics[width=0.8\textwidth]{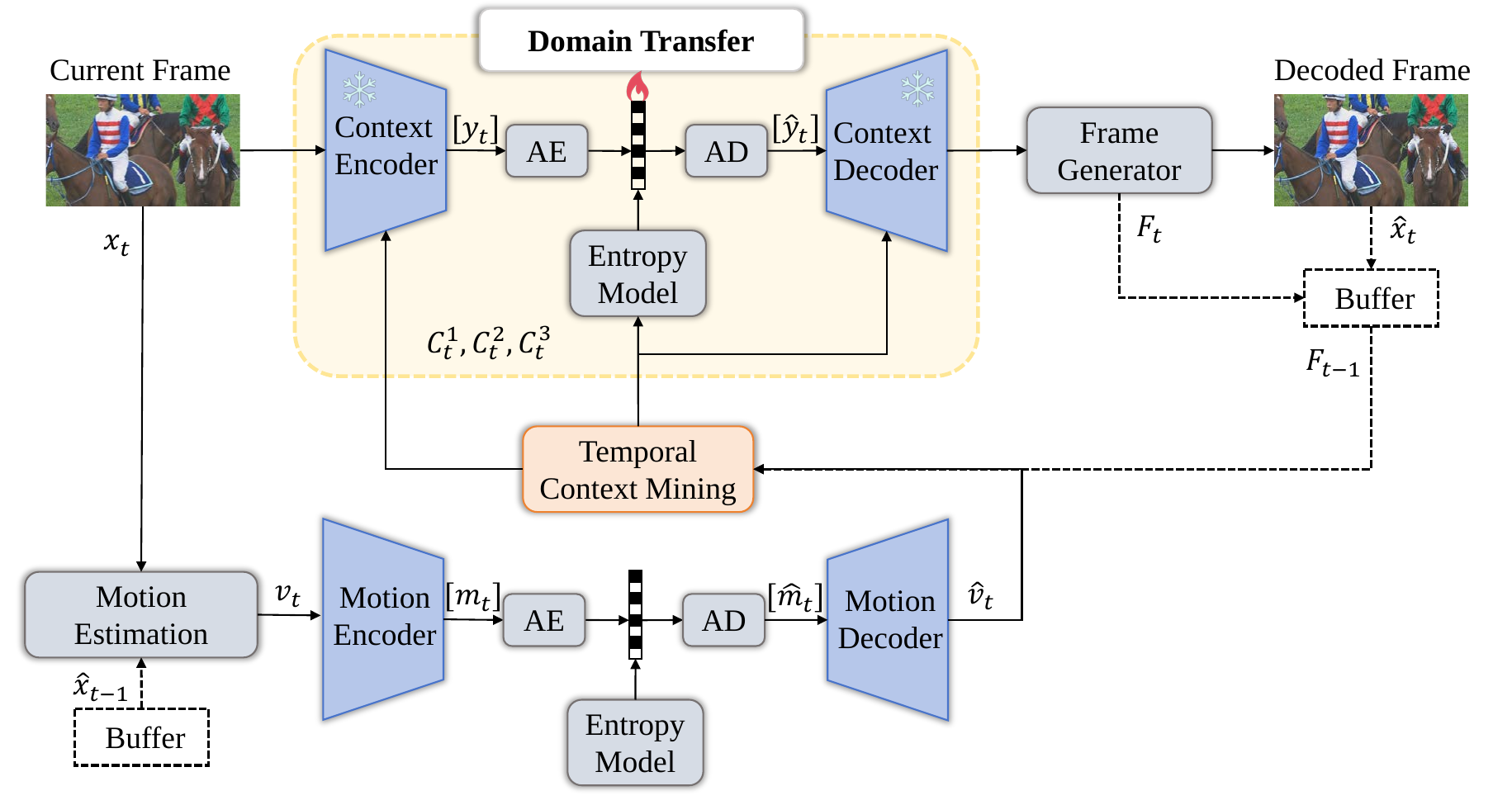}
    \caption{Overall framework of DCVC-DT. We integrate the proposed Online Domain Transfer (DT) mechanism and Dynamic RD Adjustment strategy into the baseline NVC architecture. By optimizing latent features solely on the encoder side, our framework effectively bridges the domain gap between training and testing data and enhances generalization to diverse contents.}
    \label{fig:framework}
\end{figure*}

\section{Related Works}

\subsection{Neural Video Compression}

The DCVC family~\cite{li2021deep,sheng2022temporal,li2022hybrid,li2023neural,jia2025towards,li2024neural} represents one of the most influential paradigms in neural video compression (NVC). Built upon motion estimation and compensation, DCVC~\cite{li2021deep} introduced conditional coding in the feature domain to exploit temporal contexts more effectively. DCVC-TCM~\cite{sheng2022temporal} enhanced temporal context through multi-scale feature extraction, while DCVC-HEM~\cite{li2022hybrid} improved the precision of the entropy model by jointly exploiting spatial and temporal priors. DCVC-DC~\cite{li2023neural} diversified context mining through group-based offset design, DCVC-FM~\cite{li2023neural} expanded the achievable bitrate range, and DCVC-RT~\cite{jia2025towards} explored real-time coding efficiency. These developments have established DCVC as a representative baseline for state-of-the-art NVC frameworks.

\subsection{Content-Adaptive Optimization in NVC}

Content-adaptive strategies are increasingly explored in video compression to address the domain gap between training and testing data. Tang et al.~\cite{tang2024offline} fine-tuned optical flow using motion information from VTM in an offline stage, followed by online iterative optimization to enhance motion estimation. Similarly, Chen et al.~\cite{chen2024group} extended parameter-efficient transfer learning (PETL) to NVC using low-rank adapters. Lu et al.~\cite{lu2020content} proposed an online content-adaptive framework based on rate-distortion optimization, while Zuo et al.~\cite{zuo2024frame} introduced a frame-level adaptive $\lambda$ for dynamic rate-distortion control. Additionally, Bilican et al.~\cite{bilican2025content} adjusted motion vector scales at test time to improve compression for videos with complex motion.

\section{Proposed Method}

As shown in Fig.~\ref{fig:framework}, DCVC-DT extends the condition-based NVC framework DCVC-DC~\cite{li2023neural} with two modules: an online latent refinement for domain transfer and a frame-level dynamic RD adjustment for quality fluctuation guidance. Both encoder and decoder parameters are fixed, with only the latent representation being optimized on the encoder side. This optimization does not affect decoding time and mitigates the error propagation issue present in DCVC-DC.

\subsection{Overview}

During inference, the motion estimation network generates motion vectors $v_t$ between $x_t$ and $\hat{x}_{t-1}$, which are encoded into latent motion features $m_t$ and decoded as $\hat{v}_t$. The reconstructed motion warps previous-frame features ${F}_{t-1}$ to produce multi-scale temporal contexts ${C}_t^{1,2,3}$ for conditional coding of the current frame. The latent representation $y_t$ is then quantized and entropy coded. To enhance adaptivity, DCVC-DT refines $y_t$ online through a lightweight domain transfer process and dynamic adjustment of frame-level RD based on quality variations.

\begin{figure*}[t]
    \centering
    \includegraphics[width=0.9\linewidth]{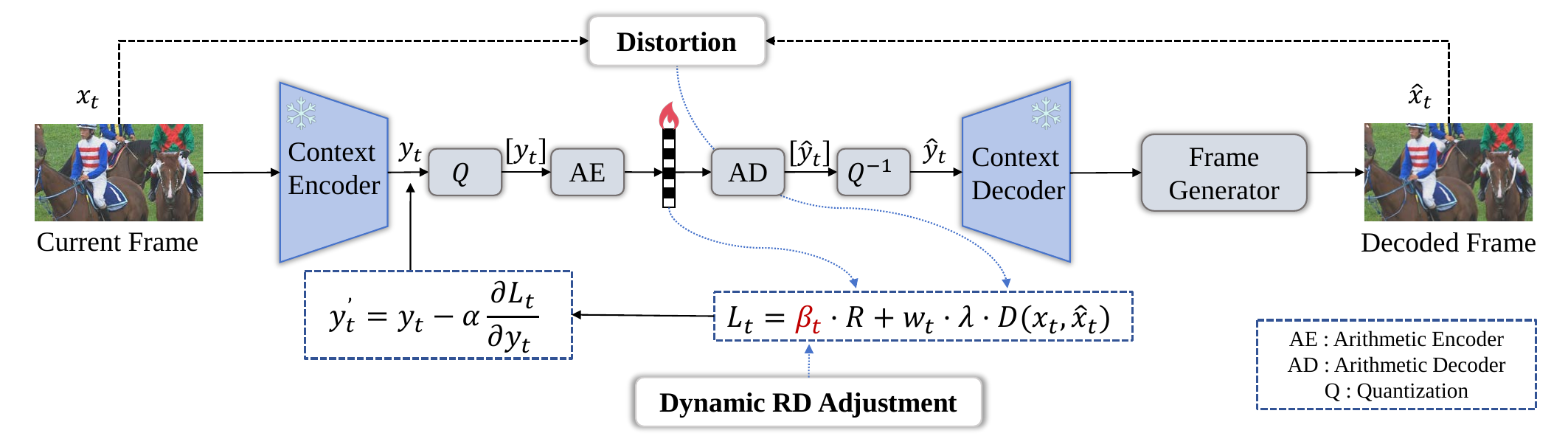}
    \caption{Illustration of the proposed Online Latent Refinement and Frame-level Dynamic RD Adjustment modules. During inference, the encoder iteratively updates the latent representation $y_t$ via gradient descent to minimize RD loss. Simultaneously, the dynamic RD adjustment module regulates the rate weight $\beta_t$ based on the inter-frame quality fluctuation $\Delta Q_t$ to effectively enhance RD performance. Notably, this entire process requires no network parameter updates.}
    \label{fig:online}
\end{figure*}

\subsection{Online Latent Refinement}

The proposed online latent refinement module operates entirely on the encoder side and refines the latent representation $y_t$ after initial encoding. Instead of updating model parameters, the module directly optimizes $y_t$ in accordance with the rate–distortion objective:
\begin{equation}
\mathcal{L}_t = R + w_t \cdot \lambda \cdot D(x_t, \hat{x}_t),
\end{equation}
where $D(\cdot)$ measures the reconstruction distortion PSNR, $R(\cdot)$ denotes the estimated bitrate, and $w_t$ is a frame-level weighting factor that controls the RD trade-off, with $w_t$ values of \{0.5, 1.2, 0.5, 0.9\} for a 4-frame miniGOP, representing factors for hierarchical quality structure adjustment.

The latent representation is updated via gradient descent:
\begin{equation}
y_t^{'} = y_t - \alpha \frac{\partial \mathcal{L}_t}{\partial y_t},
\end{equation}
where $\alpha$ is the learning rate. We replace the rounding process with stochastic Gumbel annealing based quantization (SGA-Q)~\cite{yang2020improving}, formulated as $[y_t'] = Q(y_t)$. Since only the latent representation is refined, both encoder and decoder parameters are fixed, ensuring no impact on decoding time. This lightweight process enables the encoder to adapt dynamically to domain-specific content characteristics, effectively bridging the gap between training and testing domains.

\subsection{Frame-level Dynamic RD Adjustment}

In addition to online latent refinement, DCVC-DT introduces a frame-level dynamic RD adjustment strategy to improve reconstruction quality. The method dynamically adjusts a weighting factor $\beta_t$ in the bitrate computation according to the distortion variation between consecutive frames. The updated loss function is as follows:
\begin{equation}
\mathcal{L}_t = \beta_t \cdot R + w_t \cdot \lambda \cdot D(x_t, \hat{x}_t),
\end{equation}
where $\beta_t$ is a dynamically adjusted weighting factor that modulates the bitrate $R$. The factor $\beta_t$ is based on the distortion variation between consecutive frames, and it helps to adjust the contribution of the bitrate term to the loss function, optimizing the RD trade-off for better RD performance.

Specifically, the distortion difference is computed as $\Delta Q_t = Q(x_t, \hat{x}_t) - Q(x_{t-1}, \hat{x}_{t-1})$, where $Q(\cdot)$ measures reconstruction distortion PSNR. The factor $\beta_t$ is then updated from a baseline $\beta_0 = 1$ as:
\begin{equation}
\beta_t = \beta_0 - 0.2 \cdot \text{sign}(\Delta Q_t),
\end{equation}
where $\beta_0$ is the initial value of $\beta_t$, and the adjustment step is 0.2. This value of 0.2 is empirically chosen, and experiments have shown it to be the optimal value compared to 0.1 and 0.3. It does not significantly degrade the distortion term while allowing for an increase in bitrate, which helps to improve reconstruction quality. If the current frame has a lower PSNR than the previous one, $\beta_t$ is increased by 0.2, otherwise, it is decreased by 0.2. This allows dynamic adjustment of the weights of $R$ and $D$ in the loss function based on distortion fluctuations between frames, optimizing overall RD performance.

Unlike the fixed periodic bitrate weighting schemes in DCVC-DC, where $w_t$ values are set as \{0.5, 1.2, 0.5, 0.9\}, our approach adjusts both $R$ and $D$, redistributing their contributions in the loss function. This adaptive method provides more precise control over bitrate allocation, smoothing quality fluctuations and enhancing overall rate-distortion efficiency across sequences.

\begin{table*}[t]
    \centering
    \caption{BD-rate (\%) comparison of DCVC-based methods on HEVC Class C and D datasets evaluated under PSNR. The anchor is DCVC-DC.}
    \label{tab:bd-hevc-psnr}
    \small
    \renewcommand{\arraystretch}{1.05}
    \resizebox{\textwidth}{!}{%
    \begin{tabular}{c|cccc|cccc|c} 
        \toprule
        \multirow{2}{*}{\textbf{Method}} 
        & \multicolumn{4}{c|}{\textbf{HEVC C}} 
        & \multicolumn{4}{c|}{\textbf{HEVC D}} 
        & \multirow{2}{*}{\textbf{Avg}} \\
        \cmidrule(lr){2-5}\cmidrule(lr){6-9}
        & BQMall & BasketballDrill & PartyScene & RaceHorses
        & BasketballPass & BlowingBubbles & BQSquare & RaceHorses &  \\
        \midrule
        DCVC-TCM~\cite{sheng2022temporal} & 103.99 & 101.37 & 107.23 & 77.11 & 82.13 & 81.95 & 140.53 & 66.42 & 95.09 \\
        DCVC-HEM~\cite{li2022hybrid}  & 38.85  & 54.14 & 49.09 & 30.81 & 38.84 & 40.71 & 63.29 & 33.70 & 43.68 \\
        DCVC-DC~\cite{li2023neural}    & 0.00 & 0.00 & 0.00 & 0.00 & 0.00 & 0.00 & 0.00 & 0.00 & 0.00 \\
        DCVC-DT (Ours)               & {\bf -7.95} & {\bf -5.49} & {\bf -3.22} & {\bf -8.72} 
                                     & {\bf -5.64} & {\bf -5.48} & {\bf -5.91} & {\bf -7.25} & {\bf -6.21} \\
        \bottomrule
    \end{tabular}%
    }
\end{table*}

\section{Experiments and Results}

\subsection{Experimental Setup}

We evaluate DCVC-DT on HEVC Class C and D test sequences~\cite{bossen2010common} following the evaluation settings of DCVC-DC~\cite{li2023neural}. All videos are converted to BT.709 color space for evaluation. The learning rate for the refinement process is initially set to $1 \times 10^{-3}$ and is reduced to $1 \times 10^{-4}$ after $80\%$ of the total iterations. The optimization is performed for up to $1000$ iterations for each test sequence. Different $\lambda = 8 \cdot \{85, 170, 380, 840\}$ values are applied to control the rate–distortion trade-off at various bitrate points, with the factor of 8 adjusting for the online optimization to align the results with the original bitrate. All experiments are conducted under a real encoding and decoding process with an intra period of 32. The first 96 frames are evaluated.

\subsection{Experimental Results}

\begin{figure}[t]
    \centering
    \includegraphics[width=\linewidth]{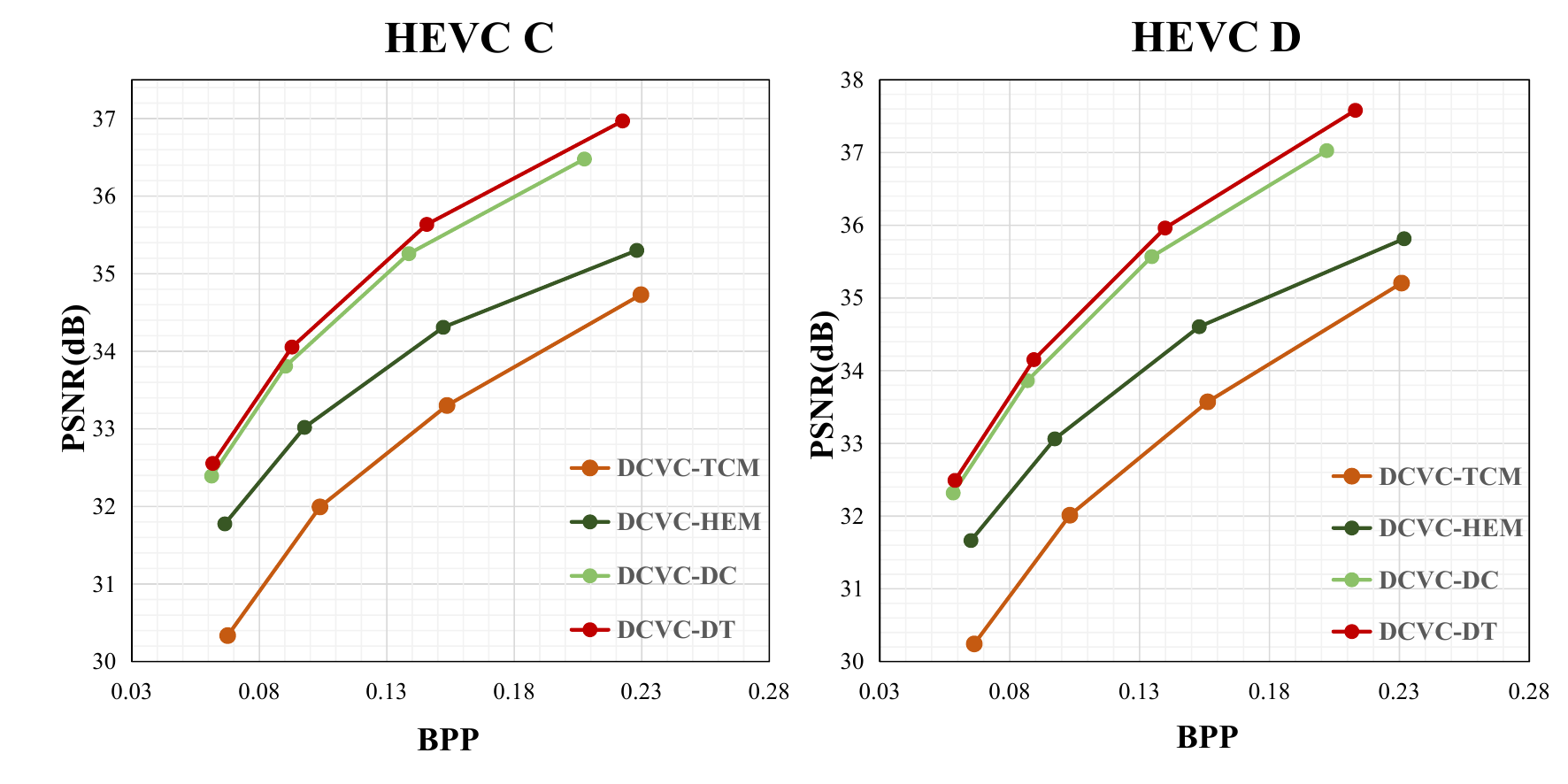}
    \caption{R–D performance comparison on HEVC Class C and D datasets evaluated with PSNR under BT.709 color space.}
    \label{fig:rdcurve}
\end{figure}

We compare DCVC-DT with representative DCVC works, including DCVC-DC~\cite{li2023neural}, DCVC-TCM~\cite{sheng2022temporal}, and DCVC-HEM~\cite{li2022hybrid}. Rate–distortion performance is evaluated in terms of PSNR on HEVC Class C and D datasets. As shown in Table~\ref{tab:bd-hevc-psnr} and Fig.~\ref{fig:rdcurve}, DCVC-DT consistently outperforms all compared methods, achieving an average BD-rate (PSNR)~\cite{bjontegaard2001calculation} reduction of 6.21\% over DCVC-DC.

\begin{figure}[htbp]
\centering
\includegraphics[width=\linewidth]{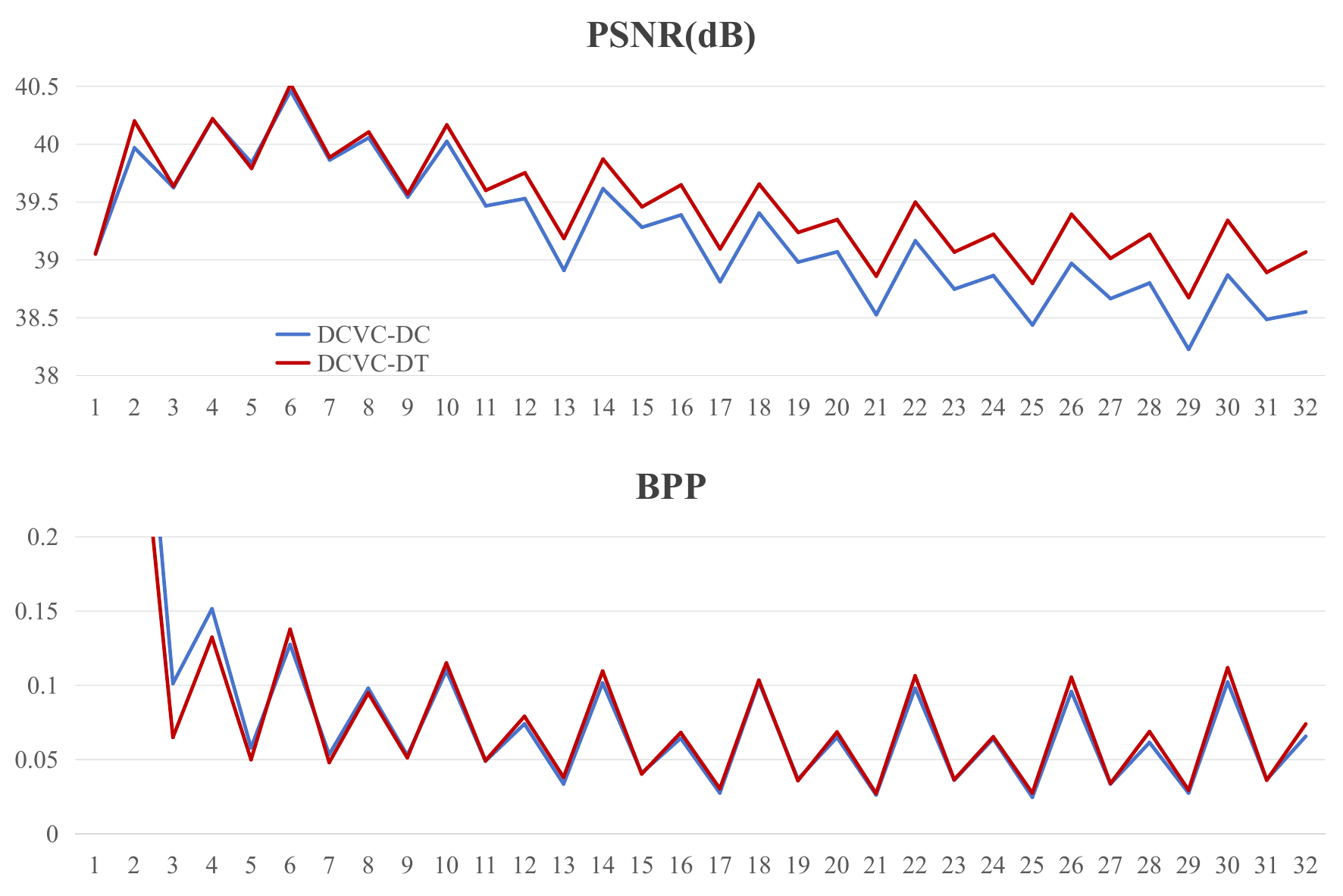}
\caption{The first 32 frames of the BasketballPass sequence from the HEVC Class D dataset, showing PSNR and BPP fluctuations at the highest bitrate. This highlights better error propagation management.}
\label{fig:QualityFluctuation}
\end{figure}

To further demonstrate the effectiveness of our method, we provide a curve visualization in Fig.~\ref{fig:QualityFluctuation}, which presents the PSNR and BPP fluctuation curves for the first 32 frames of the BasketballPass sequence from the HEVC Class D dataset at the highest bitrate. The results clearly show that DCVC-DT achieves improved reconstruction quality and significantly alleviates the error propagation issue. This is due to the continuous improvement in the quality of reference frames, which leads to progressively better performance across frames.

\subsection{Ablation Study}

To evaluate the contribution of each component, ablation experiments are conducted on first 32 frames of the \textit{BasketballPass} sequence from the HEVC Class D dataset. 
The baseline Model A corresponds to the original DCVC-DC. Model B incorporates only the online latent refinement module, while Model C further adds the frame-level dynamic RD adjustment strategy. Each refinement process is performed for 500 iterations during inference.  
As shown in Table~\ref{tab:ablation}, online latent refinement alone achieves a 2.35\% BD-rate reduction, and the integration of frame-level dynamic RD adjustment brings 3.61\% improvement. These results confirm that both modules are effective.

\begin{table}[t]
    \centering
    \caption{Ablation study on the first 32 frames of BasketballPass sequence
 from the HEVC Class D dataset with BD-rate (PSNR).}
    \label{tab:ablation}
    \renewcommand{\arraystretch}{1.1}
    \footnotesize
    \resizebox{0.48\textwidth}{!}{%
    \begin{tabular}{cccc}
        \hline
        \textbf{Component} & \textbf{Model A} & \textbf{Model B} & \textbf{Model C} \\
        \hline
        Online Latent Refinement & \xmark & \cmark & \cmark \\
        Frame-level Dynamic RD Adjustment & \xmark & \xmark & \cmark \\
        BD-Rate (\%) & 0.0 & -2.35 & -3.61 \\
        \hline
    \end{tabular}%
    }
\end{table}

\subsection{Complexity Analysis}

The proposed method performs domain transfer at the encoding side, which increases both computation and memory requirements on the encoder due to iterative optimization. If iteration is set to 500, fine-tuning a single HEVC Class D sequence at one bitrate point takes only about 25 minutes on a single A100 (40G) GPU. The number of iterations can be adjusted based on the desired level of optimization and complexity constraints. In contrast, the encoder and decoder parameters remain unchanged, and the decoding time is unaffected. This makes the method particularly suitable for scenarios requiring fine optimization on the encoder side.

\section{Conclusion}

In this work, we presented DCVC-DT, a domain transfer enhanced neural video compression framework that mitigates the domain gap between training and testing data. By introducing a lightweight online domain transfer mechanism and a frame-level dynamic RD adjustment scheme, our method achieves adaptive compression without modifying the network parameters. Experimental results demonstrate that DCVC-DT improves rate-distortion efficiency and enhances generalization across diverse testing scenarios, highlighting its strong potential for content adaptation in video encoding.

\bibliographystyle{IEEEtran}
\bibliography{refs}

\end{document}